\documentclass[letterpaper, 10 pt, conference]{ieeeconf}  

\IEEEoverridecommandlockouts                              

\usepackage{graphics} 
\usepackage{graphicx} 
\usepackage{amssymb}  
\usepackage{subfigure}
\usepackage{caption}

\usepackage{amsmath} 
\usepackage{hyperref}

\usepackage{booktabs, multirow, array, adjustbox}

\title{ \LARGE \bf Proprioceptive Image: An Image Representation of Proprioceptive Data from Quadruped Robots for Contact Estimation Learning}

\author{ Gabriel Fischer Abati$^{1,2*}$, João Carlos Virgolino Soares$^{1}$, Giulio Turrisi$^{1}$, Victor Barasuol$^{1}$, Claudio Semini$^{1}$ 
\thanks{This work was supported by the European Union – NextGenerationEU, the PNRR MUR Project PE000013 “Future Artificial Intelligence Research (FAIR)”.}
\thanks{$^{1}$ Dynamic Legged Systems (DLS) lab, Istituto
Italiano di Tecnologia (IIT), Genova, Italy.}
\thanks{$^{2}$ University of Genoa, Genoa, Italy}
\thanks{* Corresponding author, {\tt\small gabriel.fischer@iit.it}}
}

\begin{document}

\maketitle

\begin{abstract}
This paper presents a novel approach for representing proprioceptive time-series data from quadruped robots as structured two-dimensional images, enabling the use of convolutional neural networks  for learning locomotion-related tasks. The proposed method encodes temporal dynamics from multiple proprioceptive signals, such as joint positions, IMU readings, and foot velocities, while preserving the robot’s morphological structure in the spatial arrangement of the image. This transformation captures inter-signal correlations and gait-dependent patterns, providing a richer feature space than direct time-series processing. We apply this concept in the problem of contact estimation, a key capability for stable and adaptive locomotion on diverse terrains. Experimental evaluations on both real-world datasets and simulated environments show that our image-based representation consistently enhances prediction accuracy and generalization over conventional sequence-based models, underscoring the potential of cross-modal encoding strategies for robotic state learning. Our method achieves superior performance on the contact dataset, improving contact state accuracy from 87.7\% to 94.5\% over the recently proposed MI-HGNN method, using a 15 times shorter window size.

\end{abstract}

\section{INTRODUCTION}
Deep learning has been highly successful when sequential or high-dimensional signals are converted into visual representations that suit convolutional architectures. In speech recognition, raw audio is commonly transformed into spectrograms, i.e., 2D time–frequency images, enabling CNN-based models to achieve state-of-the-art results, from keyword spotting to end-to-end transcription \cite{Chan2015ListenAA, Amodei2015DeepS2, Schneider2019wav2vecUP}. Similarly,
in cybersecurity, software files and their execution behavior
have been mapped to grayscale or RGB images, where CNNs exploit visual patterns for malware detection~\cite{Llaurad2018UsingCN}. These data-to-image mappings leverage the spatial feature learning strengths of image-based deep networks.

Motivated by these successes, we transform quadruped proprioceptive time-series signals into 2D images that encode both temporal dynamics and robot morphology. Signals are arranged according to the robot’s physical layout, with time incorporated as an additional spatial dimension, preserving inter-signal correlations and gait-dependent structure in a CNN-friendly format. This yields a richer and more spatially coherent feature space than learning directly from raw time series, and we demonstrate its effectiveness on foot contact estimation, a key component for stable and adaptive locomotion \cite{Camurri2017ProbabilisticCE, Nistic2025MUSEAR}.

\begin{figure} [t!]
\centerline{\includegraphics[width=1.0\columnwidth]{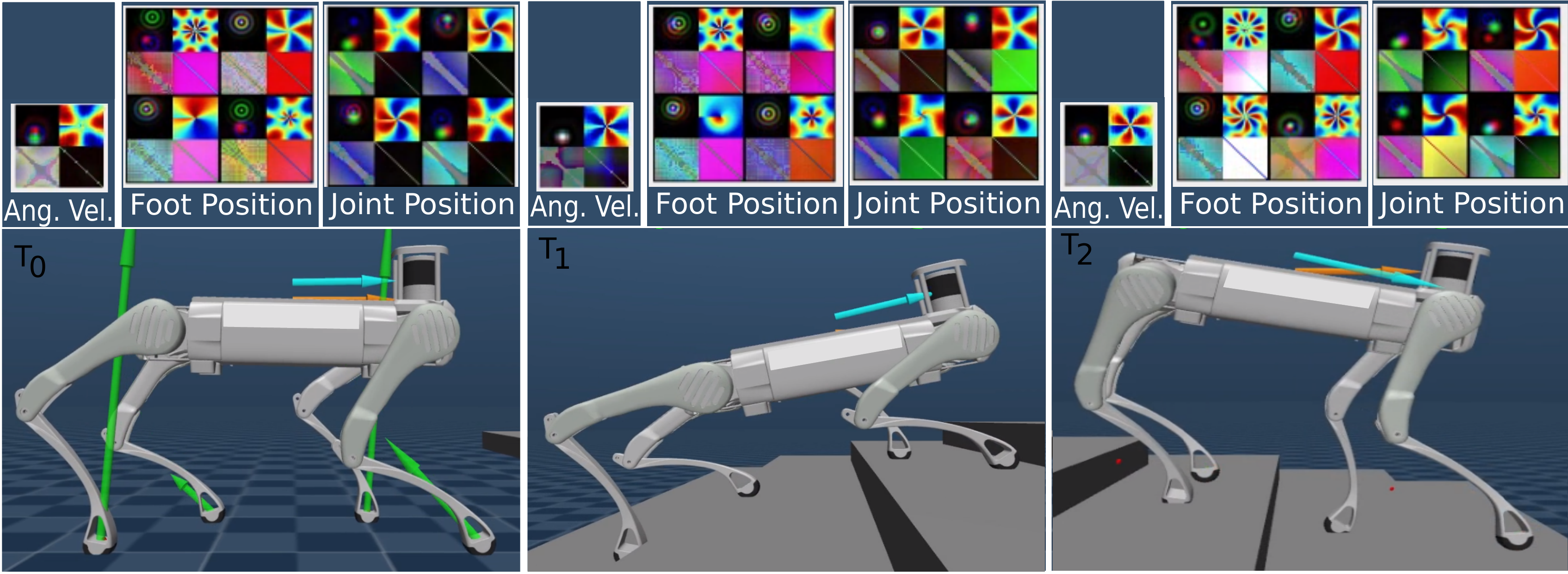}}
  \caption{
  Examples of proprioceptive images over three time steps of a quadruped robot walking on uneven terrain. Each image encodes angular velocity, joint position, and foot position, illustrating how the representation varies across environments and supports learning of tasks such as foot contact estimation.}
  \label{fig:1}
\end{figure}

Figure \ref{fig:1} shows three time steps of a quadruped traversing uneven terrain, along with the corresponding proprioceptive images derived from IMU base angular velocity, encoder joint positions, and estimated foot positions. The resulting patterns vary across environments while providing a compact representation that supports learning tasks such as contact estimation.

\subsection{Related Work - 1D time series to image conversion}

To enable CNN-based classification on time-series data, Wang and Oates \cite{Wang2014EncodingTS} proposed the Gramian Angular Field (GAF) and Markov Transition Field (MTF) encodings. GAF constructs a Gramian matrix \cite{Horn1985MatrixA} whose entries capture trigonometric relations between time indices, yielding a bijective mapping that preserves temporal correlations and supports approximate reconstruction.

In contrast, MTF represents temporal dynamics through transition statistics: the signal is discretized into quantile bins and a time-indexed transition matrix is formed \cite{Wang2014EncodingTS}. Although MTF captures multi-span transition probabilities (including self-transitions along the diagonal), it is sensitive to the binning choice, which complicates its use with heterogeneous modalities. Moreover, both MTF and GAF exhibit rate invariance, producing identical images for increasing and decreasing sequences with the same rate.

Recurrence Plots (RP) \cite{Eckmann1987RecurrencePO} encode temporal structure by marking state recurrences, supporting the diagnosis of drift, periodicity, and chaotic dynamics, but they are less responsive to subtle variations.

Overall, converting 1D time series into 2D encodings (GAF, MTF, RP) enables spatial feature learning with CNNs and has shown strong performance across benchmarks \cite{Zhang2020EncodingTS, Park2023MetaFeatureFF, Quan2023TimeSC}. Performance can be further improved via multimodal fusion (meta-features, attention, or concatenation), which captures complementary static, dynamic, and recurrence cues, with applications ranging from manufacturing defect detection to heartbeat classification \cite{Yang2019MultivariateTS, Yang2019SensorCU, Wang2015ImagingTT}. As such, fused image encodings paired with tailored CNN/attention architectures remain a practical and effective strategy for time-series classification \cite{Wang2020TimeSeriesCB, Li2022AnIF}.

In this work, we bring this paradigm to quadruped robotics by converting proprioceptive time-series signals into structured 2D representations informed by robot morphology, enabling CNN-based feature extraction for tasks such as foot contact estimation.

\subsection{Related Work - Contact Estimation}
 
 Contact estimation for quadruped robots is a fundamental perception problem, since contact measurements are not always available \cite{Nistic2025MUSEAR}. Yet it remains challenging due to sensor noise, terrain variability, and effects such as slippage.

A common baseline estimates contact by thresholding measured or estimated ground reaction forces (GRF)~\cite{Fink2020ProprioceptiveSF}, but it often fails to generalize across robots and environments. To improve robustness, several works combine probabilistic modeling, sensor fusion, and system dynamics. Camurri et al. \cite{Camurri2017ProbabilisticCE} estimate contact probabilistically using GRF computed from joint positions, velocities, and efforts. Menner and Berntorp~\cite{Menner2024SimultaneousSE}, instead, cast locomotion as a switched system, where each mode corresponds to a different set of feet in contact. Maravgakis et al. \cite{Maravgakis2023ProbabilisticCS} use Kernel Density Estimation (KDE) over IMU batches to infer the probability of stable 6-DoF contact, while the Dual $\beta$-Kalman Filter \cite{Zhang2024RobustSE} explicitly addresses slippage and variable leg length.

Despite these advances, model-based and thresholding approaches remain noise-sensitive and typically require empirical tuning, limiting robustness on lightweight robots and changing terrains. Consequently, recent work has shifted toward data-driven techniques. Youm et al.~\cite{Youm2024LeggedRS} learn a contact-probability measurement model via a Neural Measurement Network to improve IEKF performance under slip, and Lee et al. \cite{Lee2025LeggedRS} combine probabilistic torque/dynamics-based estimation with a neural compensator for residual errors. Beyond these hybrid strategies, fully learning-based methods have also emerged. For example, Lin et al.~\cite{Lin2021LeggedRS} propose Contact-CNN, a multi-modal network that relies only on proprioceptive signals (joint encoders, IMU, kinematics) to classify foot–ground contact across diverse gaits and terrains, removing the need for explicit contact sensing.

Ordonez-Apraez et al. \cite{Apraez2023OnDS} introduced Discrete Morphological Symmetries (DMSs), a group-theoretic symmetry groups of dynamical systems. Building on this, they proposed ECNN, a variant of Contact-CNN that applies sparse equivariant convolutions and batched augmentation to enforce hard equivariance, making large architectures tractable and improving leg-level F1 and test loss. Also, MI-HGNN \cite{Butterfield2024MIHGNNMH} introduces a morphology-informed heterogeneous graph neural network (GNN) that constructs a graph from a robot’s kinematic structure (nodes being the base/joints/feet, and edges representing links) to constrain learning for contact perception problems. Empirically, MI-HGNN attains substantially higher contact-detection accuracy (mean 87.7\% vs 78.8\% from ECNN \cite{Apraez2023OnDS}) on the Contact Data Set \cite{Lin2021LeggedRS}.

Although recent data-driven methods \cite{Lin2021LeggedRS, Apraez2023OnDS,Butterfield2024MIHGNNMH} have achieved promising results, they still rely on raw proprioceptive data streams as input. In contrast, our approach introduces a novel representation to create unique image patterns to describe quadruped robot sensing. To the best of our knowledge, this is the first method in robotics that converts proprioceptive signals into image representation.

\subsection{Contributions}

The contributions of the paper can be summarized as follows:

\begin{itemize}
    \item A novel representation of proprioceptive signals into 2D images, with the name Proprioceptive image (\textsc{PI}). The core objective of the \textsc{PI} representation is to encode both temporal dynamics and morphological characteristics of quadruped robots. This image representation can be explored in learned-based approaches to perform complex tasks in robotics.  
    \item We propose a CNN architecture that is able to learn from \textsc{PI} information and perform contact estimation for quadruped robots.
    \item We show the results of the \textsc{PI} trained models in real-world datasets and in simulated environments, and compared with state-of-the-art methods. In the contact dataset from \cite{Lin2021LeggedRS}, our method improved contact
state accuracy from 87.7\% to 94.5\% over the recently proposed
MI-HGNN method~\cite{Butterfield2024MIHGNNMH}, using a 15 times shorter window size. 
\end{itemize}

\section{METHODOLOGY}

\begin{figure*}
    \centerline{\includegraphics[width=1.8\columnwidth]{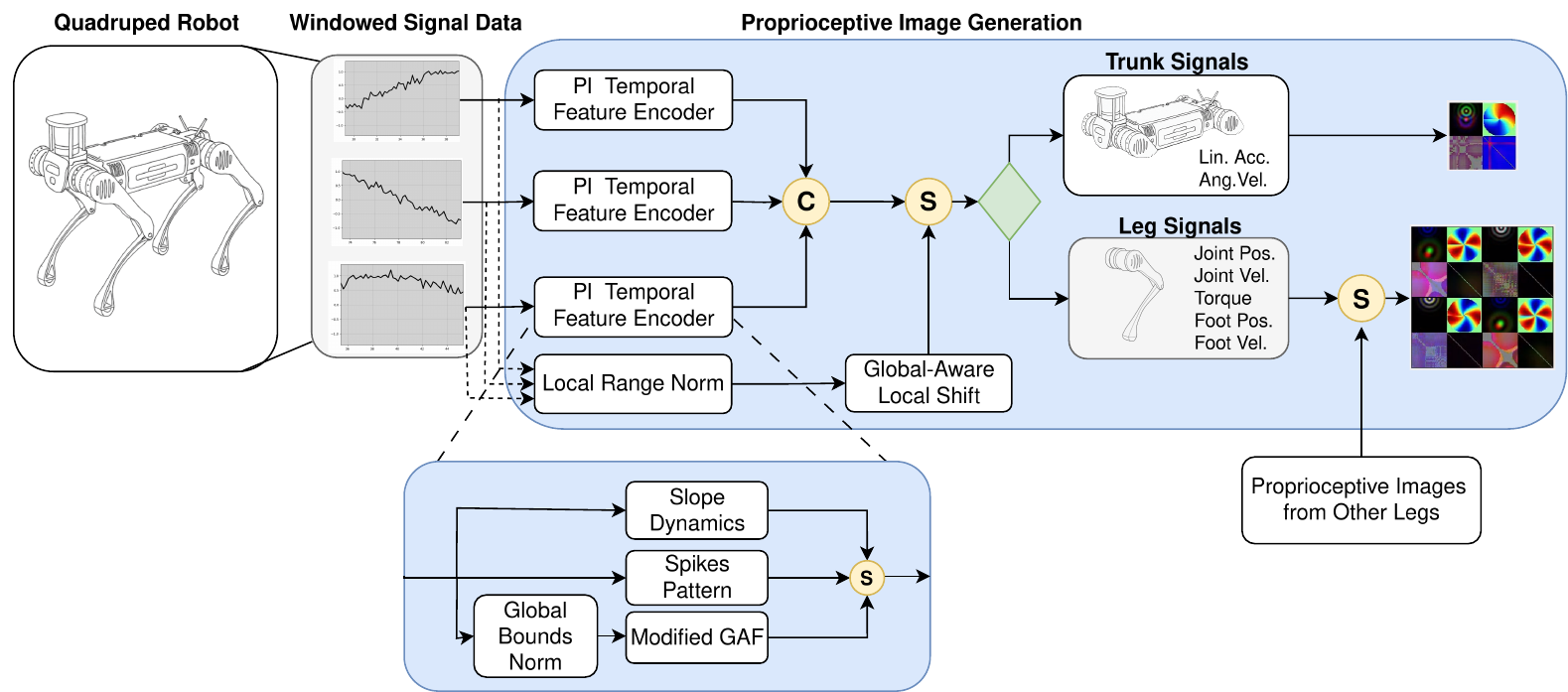}}
          \caption{ Pipeline for constructing a Proprioceptive Image. Proprioceptive signal streams are grouped in triplets (X, Y, Z for trunk signals; HAA, HFE, KFE for leg signals) to form three-channel RGB inputs. Each windowed signal is processed by a temporal feature encoder that generates image-based encodings of slope dynamics, spike patterns, and GAF. In parallel, the local range of each signal triplet is used to create the final \textsc{PI} sub-image, named Global-Aware Local Shift Image, which is spatially concatenated with the \textsc{PI} Temporal Feature Encoder into a square image. Leg \textsc{PIs} are further arranged in a $\begin{bmatrix}LF & RF \\ LH & RH \end{bmatrix}$ layout to represent all four legs, while trunk \textsc{PIs} are used individually. Yellow circles labeled `C' indicate channel concatenation, and circles labeled `S' denote spatial concatenation.}
  	\label{fig:pimeth}
\end{figure*}

 In this section, we present the four key components of the \textsc{PI} formulation and explain how they are fused to construct the final \textsc{PI} representation, which is tailored to the robot’s morphology. Each component captures essential temporal and statistical characteristics of a single proprioceptive state and encodes them into a sub-image through well-defined mathematical transformations. Specifically, it is extracted and encoded into image form using the following elements from the signal data stream: the slope dynamics in Section~\ref{met:slope_dyn}, spike patterns in Section~\ref{met:spikes}, modified GAF polar distance in Section~\ref{met:polar_distance}, and the global-aware local shift in Section~\ref{met:local_shift}. Figure \ref{fig:pimeth} shows how to generate the Proprioceptive Image from proprioceptive signal data.
 
\subsection{Preliminaries}
\textbf{Input Values}: Let $s = \{s_i\}^T_{i=1}$ be a discrete 1D time series sampled from a proprioceptive signal (estimated or measured by a sensor). At each timestamp $t$, we define a sliding window of length $w$ as $x_t = [s_{t-w+1}, \dots, s_t] \in \mathbb{R}^w$. The goal is to convert each key component into a $w\times w$ image and concatenate them to form a $ 2w \times 2w $ image.

\textbf{Global Bounds}: The \textsc{PI}'s representation uses as reference the global bounds of the input signal. The global bounds are defined as the maximum and minimum values of the proprioceptive signals outputs $\mathcal{G} = [s^{global}_{min}, s^{global}_{max}]$ (usually obtained from the sensor datasheet). $\mu_{\mathcal{G}}$ is defined as the average of the global bounds $\mu_{\mathcal{G}} = (s^{global}_{max} + s^{global}_{min})/2$. The normalized sliding window sequence, computed with respect to the global bounds, is defined as 
$x_t^{\mathcal{G}} = [s_{t-w+1}^{\mathcal{G}}, \dots, s_t^{\mathcal{G}}] \in \mathbb{R}^w$, where each $s_i^{\mathcal{G}} \in [-1,1]$ represents the globally normalized signal value at time $i$ computed as:
\begin{equation}
\label{eq:normalization}
    s_i^{\mathcal{G}} = \frac{(s_i -s^{global}_{max})+(s_i - s^{global}_{min})}{(s^{global}_{max} - s^{global}_{min})}
\end{equation}
\textbf{Local Range}: For each proprioceptive signal of interest, we compute a pair of normalized deviation measures $\delta$. These values quantify how far the current local operating range $\mathcal{Y}=[s^{local}_{min},s^{local}_{max}]$ of the signal is shifted relative to its global reference range. The local range is estimated from a sliding window by taking the lower and upper percentiles (10th and 90th) and expanding them by a small margin to ensure robustness to noise. Given the global average $\mu_{\mathcal{G}}$ and global bounds $\mathcal{G}$, the normalized deviation $\delta = [\delta_{min}, \delta_{max}]$ is defined as:
\begin{equation}
\begin{array}{l@{\quad\quad}l}
  \delta_{min}=\frac{s^{local}_{min} - \mu_{\mathcal{G}}}{|\mu_{\mathcal{G}}-s^{global}_{min}|} 
  & 
  \delta_{max}=\frac{s^{local}_{max} - \mu_{\mathcal{G}}}{|\mu_{\mathcal{G}}-s^{global}_{max}|}
\end{array}
\end{equation}

\textbf{Quadruped Robot Morphology}: The quadruped robot consists of a rigid trunk and four identical legs — left-front (LF), right-front (RF), left-hind (LH), and right-hind (RH). Each leg has three actuated joints: hip abduction/adduction (HAA), hip flexion/extension (HFE), and knee flexion/extension (KFE). Each joint provides proprioceptive feedback, including torque (or torque estimates from motor currents), angular position, and velocity. The trunk houses an inertial measurement unit (IMU) for measuring accelerations, angular velocities, and estimating body orientation.

\textbf{\textsc{PI} Concatenation}:
The primary objective of each \textsc{PI} is to capture key dynamics of a data stream and encode them into a compact image representation. To also incorporate the robot’s morphology, individual \textsc{PIs} are concatenated both channel-wise and spatially to form a final representation for a given proprioceptive signal type (e.g., torque, joint position, joint velocity). \textsc{PIs} are grouped into leg \textsc{PIs} and trunk \textsc{PIs}. Leg signals include per-leg proprioceptive signals such as joint positions, joint velocities, foot positions, foot velocities, and GRFs. Trunk signals, such as the data from IMUs, provide measurements of the trunk’s motion in space (angular velocity and linear acceleration), which serve as global indicators of the robot’s state. For example, the torque \textsc{PI} for a single leg $\textsc{PI}^{Leg} \in \mathbb{R}^{2w \times 2w \times 3}$ is formed by concatenating channel-wise the \textsc{PIs} of its three joints (or three linear dimensions in the case of foot positions and foot velocities):
\begin{equation}
    \textsc{PI}^{Leg}_{Torque} = \begin{bmatrix} (\textsc{PI}^{\text{HAA}}_{Torque},\textsc{PI}^{\text{HFE}}_{Torque},\textsc{PI}^{\text{KFE}}_{Torque})
    \end{bmatrix}
\end{equation}
and the full-body torque \textsc{PI} is obtained by arranging the per-leg torque \textsc{PIs} spatially:
\begin{equation}
    \textsc{PI}_{\text{Torque}} = 
        \begin{bmatrix}
        \textsc{PI}^{LF}_{Torque} & \textsc{PI}^{RF}_{Torque} \\
        \textsc{PI}^{LH}_{Torque} & \textsc{PI}^{RH}_{Torque}
        \end{bmatrix}
        \in \mathbb{R}^{4w \times 4w \times 3}
\end{equation}

Trunk \textsc{PIs} are constructed by concatenating the components of each dimension channel-wise. The concatenation technique is used in all key components of the \textsc{PI} except the Global-Aware Local Shift.

\subsection{Slope Dynamics}
\label{met:slope_dyn}
Given the temporal window of scalar values $x_t$, we encode three dynamical descriptors — mean slope, jerk, and peak density — into a spatial Gaussian blob modulated by concentric ripples. More precisely, we fit a linear model $x_i = \beta_t i + b$ over the window $w$ to compute the slope $\beta_t \in \mathbb{R}$ using least-squares and clipped to be $[-1,1]$:
\begin{equation}
    \hat{\beta}_t = \text{clip} \Big( \frac{\beta_t}{\sigma(x) + \epsilon}, -1, 1\Big)
\end{equation}
where $\sigma(\cdot)$ is the standard deviation of the window, and $\epsilon$ ensures numerical stability. The normalized slope $\hat{\beta}$ determines the horizontal displacement of the blob. The jerk is computed as the absolute value of the second time-derivative,
\begin{equation}
  \mathcal{J} = \frac{1}{w-2} \sum_{t=0}^{w-3} |(x_{t+2}-x_{t+1})-(x_{t+1}-x_t)|
\end{equation}

\noindent and normalized as $\hat{\mathcal{J}}=\text{clip}(\frac{\mathcal{J}}{\sigma(\dot{x}_t) + \epsilon},0,1)$ by the velocity deviation $\sigma(\dot x)$, which defines the vertical displacement of the blob. The peak density is defined as the ratio between the number of local maxima and the window size, scaled to $[0,1]$:
\begin{equation}
    \text{peak} = \text{clip}\Big(\frac{\#peaks}{w} \cdot 5,0,1 \Big)
\end{equation}
which controls the ripple frequency $f_{ripple}=2+ \lfloor 6 \cdot \text{peak} \rfloor$. These normalized features are mapped to the center of a Gaussian blob, whose coordinates $c_x$ and $c_y$ are given by:
\begin{equation}
\begin{array}{l@{\quad\quad}l}
  c_x = \lfloor \frac{\hat{\beta}+1}{2} (w-1) \rfloor 
  & 
  c_y = \lfloor (1 - \hat{\mathcal{J}})(w -1) \rfloor
\end{array}
\end{equation}

\begin{figure}[h!]
\centering
  \subfigure[]{%
    \includegraphics[width=.25\linewidth]{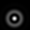}
    \label{fig:slope1}
  } 
  \subfigure[]{%
    \includegraphics[width=.25\linewidth]{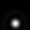}
    \label{fig:slope2} 
  }
  \subfigure[]{%
    \includegraphics[width=.25\linewidth]{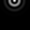}
    \label{fig:slope3}
  } 
  \caption{Examples of Slope Dynamics images from simulated joint position signal of a quadruped robot trotting.} 
  \label{fig:slopeDyn}
\end{figure}

Let $R_{i,j} = \sqrt{(i -c_x)^2 + (j - c_y)^2}$ be the radial distance from the blob center, the base Gaussian is computed as:
\begin{equation}
G_{i,j} = exp{\Bigg(- \frac{R_{i,j}^2}{2 \sigma_g^2}\Bigg)}
\end{equation}
with Gaussian spread of $\sigma_g = w/6.25$. To introduce the ripple texture, we apply a cosine modulation:
\begin{equation}
    \rho_{i,j} = 0.5 + 0.5  \cos\Bigg(\frac{f_{ripple}R_{i,j}}{\sigma_g}\Bigg)
\end{equation}
and form the final pattern $A=G \odot \rho$, normalized to the interval $[0,1]$ and scaled to 8-bit intensity. In this representation, the horizontal position of the blob encodes trend direction (slope), the vertical position encodes motion irregularity (jerk), and the ripple density reflects oscillatory content (peak density), providing a compact visual signature of local signal dynamics. Figure~\ref{fig:slopeDyn} presents three example images generated from different input signals.

\subsection{Spikes Patterns}
\label{met:spikes}
Given a temporal window $x_t$ from the proprioceptive sensor, we form the pairwise difference matrix $\Delta_{i,j} = x_j - x_i \in \mathbb{R}^{w \times w}$  and suppress noise via an adaptive gating threshold based on robust variability. Specifically, we compute the median $m_x=\text{median}(x)$ and the median absolute deviation (MAD)
\begin{equation}
    MAD_x = \text{median}(|x_t - m_x|) + \epsilon
\end{equation}
and set the gating threshold as $th_{\text{MAD}}= \alpha \text{MAD}_x$, where $\alpha > 0$ controls the sensitivity (smaller values detect more subtle changes, and larger values restrict detection to stronger transitions). A binary gate
\begin{equation}
    \mathcal{M}_{i,j} = \mathbb{I}(|\Delta_{i,j}| > th_{MAD_x})
\end{equation}
selects only pairs exceeding this adaptive threshold. For the masked upper-triangular entries $(i < j, \mathcal{M}_{i,j}=1)$, we compute their median $m_{\Delta}$ and $ \text{MAD}_{\Delta}$ to form the following robust score
\begin{equation}
Z_{i,j} = \frac{\Delta_{ij}- m_\Delta}{MAD_{\Delta}}
\end{equation}

\begin{figure}[h!]
\centering
  \subfigure[]{%
    \includegraphics[width=.25\linewidth]{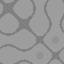}
    \label{fig:spike1}
  } 
  \subfigure[]{%
    \includegraphics[width=.25\linewidth]{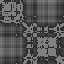}
    \label{fig:spike2} 
  }
  \subfigure[]{%
    \includegraphics[width=.25\linewidth]{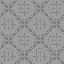}
    \label{fig:spike3}
  } 
  \caption{Examples of Spikes pattern images from simulated joint position of a quadruped robot trotting.} 
  \label{fig:Spikes}
\end{figure}

Scores are symmetrically clipped to [-3,3], following the three-sigma principle \cite{Brill2004AppliedSA} to limit extreme outliers while maintaining contrast for typical values. The final spike image $I \in [0,255]^{w \times w}$ is obtained as
\begin{equation}
    C_{i,j} = 
    \begin{cases}
    128 + 127 \frac{\text{clip}(Z_{ij},-3,3)}{3} & \mathcal{M}_{ij} = 1 \\
    128 & \mathcal{M}_{ij}=0
    \end{cases}
\end{equation}
where bright pixels $(\approx 255)$ represent strong positive spikes, dark pixels $(\approx 0)$ strong negative spikes, and mid-gray $(\approx 128)$ either neutral changes or subthreshold differences. This adaptive MAD-based construction yields a symmetric, noise-resilient spike matrix that adjusts to local variability while preserving sensitivity to significant temporal transitions. Figure \ref{fig:Spikes} illustrates three examples of proprioceptive images obtained from distinct input sequences.

\subsection{Polar Distance}
\label{met:polar_distance}
\begin{figure}[h!]
\centering
  \subfigure[]{%
    \includegraphics[width=.25\linewidth]{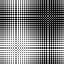}
    \label{fig:gaf1}
  } 
  \subfigure[]{%
    \includegraphics[width=.25\linewidth]{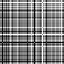}
    \label{fig:gaf2} 
  }
  \subfigure[]{%
    \includegraphics[width=.25\linewidth]{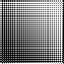}
    \label{fig:gaf3}
  } 
  \caption{Example of GAF images from simulated joint position signal of a quadruped robot trotting.} 
  \label{fig:GAF}
\end{figure}

In this sub-image, we employ a modified version of the GAF representation. GAF maps the input signal to an angular coordinate $\phi_t = arccos(x^{\mathcal{G}}_t)$ and its matrix entries are defined as $GAF=\cos(\phi_i + \phi_j), (i,j)=1,\dots,w$. While the standard GAF inherently captures static temporal information through its unique polar encoding, it suffers from ambiguity in distinguishing between increasing and decreasing trends. To overcome this limitation, we replace the values along the main diagonal of the GAF matrix with the globally normalized sliding window sequence $x_t^{\mathcal{G}}$, subsequently scaled to the pixel intensity range [0, 255]. This substitution embeds directional information directly into the matrix diagonal values, enhancing the interpretability of the encoded temporal dynamics. Figure \ref{fig:GAF} shows three representative proprioceptive images corresponding to different input conditions. Eq. \ref{eq:gaf} describes the GAF function: 

\begin{equation}
\centering
    D = \begin{bmatrix}
    x^{\mathcal{G}}_1 & \cos(\phi_{1} + \phi_2) & \dots & \cos(\phi_1 + \phi_w) \\
    \cos(\phi_2 + \phi_1) & x^{\mathcal{G}}_2 & \dots & \vdots \\
    \vdots & \vdots & \ddots & \vdots \\
    \cos(\phi_w + \phi_1) & \dots & \dots & x_w^{\mathcal{G}}
    \end{bmatrix}
\label{eq:gaf}
\end{equation}

\subsection{Global-Aware Local Shift}
\label{met:local_shift}
We propose the cymatic encoding, inspired by physical vibration patterns observed on resonating plates and membranes, where standing wave interference produces stable, symmetric geometric patterns that depend on the excitation frequencies and phases. Such patterns are a natural 2D analog of harmonic decomposition in 1D signals: radial and angular wave modes act like spatial Fourier components, while their interference encodes complex multi-frequency relationships in a compact visual form.

This sub-image is the only one that does not follow the \textsc{PI} concatenation method as the other three sub-images. Instead of inputting the local temporal window and concatenating the \textsc{PI} images for each joint/axis, we input the normalized deviations $\delta$ of three joint/axis proprioceptive signals as a 6-dimensional descriptor vector that controls the spatial pattern parameters. The first four components set the radial frequency $k_r \in (1,4)$, angular frequency $k_t \in (1,4)$, and their respective phases $\phi_r$,$\phi_t \in [0,2\pi)$. These determine the number of concentric rings and azimuthal lobes, directly modulating spatial symmetry. The fifth component sets a modulation scale $\alpha \in [0,1]$, controlling the amplitude of a secondary interference field, while the sixth is a blend weight, $m \in [0,1]$ determining how strongly the primary and secondary fields combine.

The 2D spatial domain $(X,Y)\in [-1,1]^2$ is converted to polar coordinates
\begin{equation}
\begin{array}{l@{\quad\quad}l}
 \mathcal{R} = \text{clip}(\sqrt{X^2 + Y^2},0,1)&
 T=\text{atan2}(Y,X)
\end{array}
\end{equation}
and two wave-interference components are defined as $\mathcal{C}_1 = \sin(k_r \mathcal{R} + \phi_r) \cos(k_t T + \phi_t)$ and $\mathcal{C}_2 = \sin(k_r \mathcal{R} - k_t T)$.

\begin{figure}[h!]
\centering
  \subfigure[]{%
    \includegraphics[width=.25\linewidth]{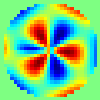}
    \label{fig:cymatic1}
  } 
  \subfigure[]{%
    \includegraphics[width=.25\linewidth]{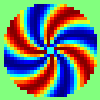}
    \label{fig:cymtic2} 
  }
  \subfigure[]{%
    \includegraphics[width=.25\linewidth]{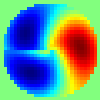}
    \label{fig:cymatic3}
  } 
  \caption{Examples of Global-Aware Local Shift images from simulated joint position signal of a quadruped robot trotting.} 
  \label{fig:cymatic}
\end{figure}

The final cymatic field is a convex combination given as
\begin{equation}
    \mathcal{C} = (1 -m) \mathcal{C}_1 + m \alpha \mathcal{C}_2
\end{equation}
masked outside the unit disk and normalized to [0, 255] for image representation. 
This mapping is special since it translates a low-dimensional descriptor into a high-complexity, structured, and interpretable 2D pattern, whose symmetry, periodicity, and interference texture vary smoothly with its parameters. Unlike arbitrary texture generators, cymatic patterns are rooted in physical wave phenomena, yielding repeatable geometric signatures that are both parameter-sensitive and visually distinct. This makes them particularly well-suited for encoding multi-feature signal descriptors into images for downstream convolutional processing, as small variations in the descriptor lead to smooth but discriminative changes in the spatial pattern. Figure \ref{fig:cymatic} depicts three sample images illustrating how varying inputs produce distinct representations.

\subsection{Contact Estimation}
\label{met:contact_estimation}
\begin{figure}[h!]
    \centerline{\includegraphics[width=1.0\columnwidth]{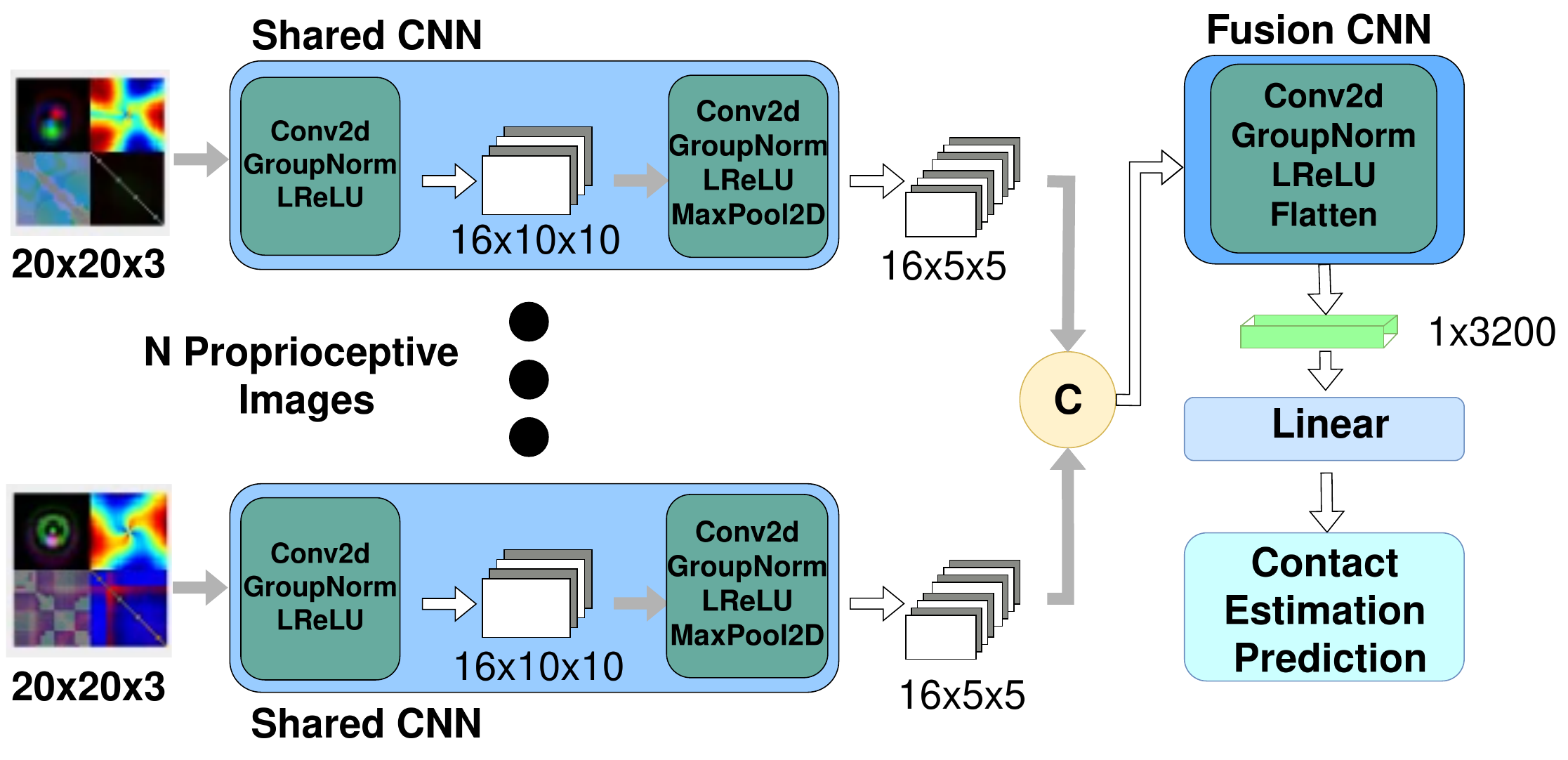}}
          \caption{Architecture of ConcatCNN for quadruped contact estimation. Each proprioceptive image is encoded by a shared CNN, and the resulting features are stacked and fused to jointly learn from multiple proprioceptive signals. In the case of fusing trunk and leg \textsc{PIs}, the concatenated four legs \textsc{PI} are separated. The fused representation is flattened and mapped through fully connected layers to predict discrete foot–ground contact states as a classification label.}
  	\label{fig:contact_estimation}
 \end{figure}

Following \cite{Lin2021LeggedRS}, contact estimation is framed as a multi-class classification problem, where each binary foot–ground configuration corresponds to a discrete class. To address this, we propose ConcatCNN, a multi-image convolutional architecture that jointly encodes and fuses proprioceptive image representations. The input modalities are joint positions (rad), joint velocities (rad/s), base linear acceleration ($m/s^2$), base angular velocity (rad/s), foot positions (m), and foot velocities (m/s), all expressed in the base frame when applicable. Each input image is processed by a shared CNN encoder with two convolutional layers (kernel size $3\times3$, padding 1), each followed by Group Normalization (GN) \cite{Wu2018GroupN} and LReLU activation \cite{Maas2013RectifierNI}, with a $2\times2$ max-pooling operation for downsampling. LReLU mitigates the risk of dead neurons in sparse proprioceptive inputs, while ReLU activations are retained in the fully connected layers for stability and efficiency. GN was chosen over Batch Normalization for its robustness to small batch sizes and heterogeneous signals streams, ensuring consistent feature scaling. The shared encoder outputs feature maps of size $16 \times 5 \times 5$ per image, which are concatenated into a fused representation of size $16N \times 5 \times 5$, where $N$ denotes the number of PIs provided as input. This representation is refined by a convolutional fusion layer (kernel size $3\times3$, 32 channels), flattened, and passed through a two-layer fully connected network (hidden dimension 128, ReLU activation) with dropout \cite{Hinton2012ImprovingNN} for regularization. The final output layer predicts the contact state across the predefined classes.
Overall, ConcatCNN enables efficient joint feature extraction and fusion by leveraging parameter sharing in the encoder and channel-wise concatenation for cross-signal integration, thereby enhancing the robustness of contact state classification. The complete architecture is depicted in Fig.~\ref{fig:contact_estimation}, which highlights the shared encoders, fusion stage, and final prediction layers.

\section{RESULTS}
\subsection{Real-World Dataset Results}
\label{sec:real_world}
In \cite{Lin2021LeggedRS}, a comprehensive open-source contact dataset was developed using the MIT Mini Cheetah robot, comprising approximately one million labeled data points collected over eight different terrains (asphalt, concrete, forest, grass, middle pebble, small pebble, rock road, and sidewalk) and multiple gaits. This dataset incorporates both contact and non-contact scenarios to enhance generalization. Ground truth contact labels were generated with a self-supervised algorithm that analyzes filtered foot height signals, thereby circumventing the need for additional external sensors during data collection.

The dataset was converted into \textsc{PI} representations using a sliding window size of $w=10$, generating images $I \in \mathbb{R}^{20 \times20 \times 3}$. The proposed contact estimation model, described in Section~\ref{met:contact_estimation}, was trained with the same proprioceptive signal modalities as \cite{Lin2021LeggedRS} to ensure a fair comparison. These modalities include joint positions, joint velocities, foot positions, foot velocities, trunk angular velocity, and trunk linear acceleration. Performance was assessed in terms of overall accuracy, F1 score, per-leg accuracy, and LegAvg accuracy (the mean of the individual leg accuracies). Table~\ref{tab:multi_dataset_results} summarizes the results across the different test sets, comparing our \textsc{PI}-based model against the approach in \cite{Lin2021LeggedRS}. The \textsc{PI} model achieved consistently higher overall accuracy and F1 score, with the latter balancing precision and recall to account for potential class imbalance, while per-leg and LegAvg accuracies were nearly identical across methods, differing only marginally. This indicates that \textsc{PI} representations yield more robust global classification performance while preserving per-leg accuracy at a high level, despite relying on sequences of only 10 time steps compared to 150 in \cite{Lin2021LeggedRS}.

\begin{table}[h!]
\centering
\begin{adjustbox}{width=0.48\textwidth}
\begin{tabular}{l c c c c c c c c}
\toprule
\textbf{\LARGE Dataset} & \textbf{\LARGE Method} & \textbf{\LARGE Acc(\%)} & \textbf{\LARGE F1} & \textbf{\LARGE LF(\%)} & \textbf{\LARGE RH(\%)} & \textbf{\LARGE RF(\%)} & \textbf{\LARGE LH(\%)} & \textbf{\LARGE Leg Avg(\%)} \\
\midrule
\multirow{2}{*}{\textbf{\LARGE Concrete}} 
& \LARGE \textsc{PI} CNN     & \textbf{ \LARGE 94.05} & \textbf{\LARGE 0.9337} & \LARGE 98.19 & \LARGE 97.33 & \LARGE 98.03 & \LARGE 98.32 & \LARGE 97.97 \\
\cmidrule(lr){2-9}
& \LARGE Lin \textit{et al.} \cite{Lin2021LeggedRS} & \LARGE 93.04  & \LARGE 0.4145 & \LARGE 97.42 & \LARGE 97.12 & \LARGE 97.49 & \LARGE 97.58 & \LARGE 97.40 \\
\midrule
\multirow{2}{*}{\textbf{\LARGE Forest}} 
& \LARGE \textsc{PI} CNN     & \textbf{\LARGE 91.09} & \textbf{\LARGE 0.8980} & \LARGE 96.49 & \LARGE 97.66 & \LARGE 96.89 & \LARGE 97.07 & \LARGE 97.03 \\
\cmidrule(lr){2-9}
& \LARGE Lin \textit{et al.} \cite{Lin2021LeggedRS}  & \LARGE 89.95     & \LARGE 0.5416    & \LARGE 96.39 & \LARGE 97.54 & \LARGE 97.11 & \LARGE 96.81 & \LARGE 96.96 \\
\midrule
\multirow{2}{*}{\textbf{\LARGE Grass}} 
& \LARGE \textsc{PI} CNN  & \textbf{\LARGE 94.27} & \textbf{\LARGE 0.9333} & \LARGE 98.21 & \LARGE 98.08 & \LARGE 98.11 & \LARGE 98.17 & \LARGE 98.14 \\
\cmidrule(lr){2-9}
& \LARGE Lin \textit{et al.} \cite{Lin2021LeggedRS}  & \LARGE 92.87 & \LARGE 0.5348 & \LARGE 97.58 & \LARGE 97.31 & \LARGE 97.85 & \LARGE 97.45 & \LARGE 97.55 \\
\midrule
\multirow{2}{*}{\textbf{\LARGE Asphalt Road}} 
& \LARGE \textsc{PI} CNN  & \textbf{\LARGE 95.20} & \textbf{\LARGE 0.9520} & \LARGE 98.63 & \LARGE 98.42 & \LARGE 98.71 & \LARGE 98.88 & \LARGE 98.66 \\
\cmidrule(lr){2-9}
& \LARGE Lin \textit{et al.} \cite{Lin2021LeggedRS}  & \LARGE 94.13     & \LARGE 0.6023 & \LARGE 97.74 & \LARGE 98.19 & \LARGE 98.46 & \LARGE 98.28 & \LARGE 98.17 \\
\midrule
\multirow{2}{*}{\textbf{\LARGE Middle Pebble}} 
& \LARGE \textsc{PI} CNN  & \textbf{\LARGE 93.18} & \textbf{\LARGE 0.9270} & \LARGE 97.77 & \LARGE 97.49 & \LARGE 97.54 & \LARGE 97.70 & \LARGE 97.62 \\
\cmidrule(lr){2-9}
& \LARGE Lin \textit{et al.} \cite{Lin2021LeggedRS}  & \LARGE 92.45     & \LARGE 0.4861    & \LARGE 97.84 & \LARGE 96.90 & \LARGE 97.16 & \LARGE 97.94 & \LARGE 97.46 \\
\midrule
\multirow{2}{*}{\textbf{\LARGE Small Pebble}} 
& \LARGE \textsc{PI} CNN  & \textbf{\LARGE 91.80} & \textbf{\LARGE 0.9032} & \LARGE 97.82 & \LARGE 96.11 & \LARGE 97.94 & \LARGE 97.62 & \LARGE 97.37 \\
\cmidrule(lr){2-9}
& \LARGE Lin \textit{et al.} \cite{Lin2021LeggedRS}  & \LARGE 90.65  & \LARGE 0.4953 & \LARGE 96.88 & \LARGE 96.41 & \LARGE 97.85 & \LARGE 97.27 & \LARGE 97.10 \\
\midrule
\multirow{2}{*}{\textbf{\LARGE Sidewalk}} 
& \LARGE \textsc{PI} CNN  & \textbf{\LARGE 95.21} & \textbf{\LARGE 0.9446} & \LARGE 98.18 & \LARGE 97.61 & \LARGE 98.20 & \LARGE 98.49 & \LARGE 98.12 \\
\cmidrule(lr){2-9}
& \LARGE Lin \textit{et al.} \cite{Lin2021LeggedRS}  & \LARGE 93.71  & \LARGE 0.4174 & \LARGE 97.43 & \LARGE 97.68 & \LARGE 97.65 & \LARGE 97.06 & \LARGE 97.53 \\
\midrule
\multirow{2}{*}{\textbf{\LARGE Rock Road}} 
& \LARGE \textsc{PI} CNN  & \textbf{\LARGE 93.72} & \textbf{\LARGE 0.9313} & \LARGE 97.69 & \LARGE 98.57 & \LARGE 97.39 & \LARGE 98.79 & \LARGE 98.11 \\
\cmidrule(lr){2-9}
& \LARGE Lin \textit{et al.} \cite{Lin2021LeggedRS}  & \LARGE 92.32 & \LARGE 0.5338 & \LARGE 97.32 & \LARGE 97.67 & \LARGE 97.29 & \LARGE 97.55 & \LARGE 97.46 \\
\bottomrule
\end{tabular}
\end{adjustbox}
\caption{Contact Estimation comparison across datasets from \cite{Lin2021LeggedRS}.}
\label{tab:multi_dataset_results}
\end{table}

\begin{figure}[h!]
  \centering
  \includegraphics[width=1.0\linewidth]{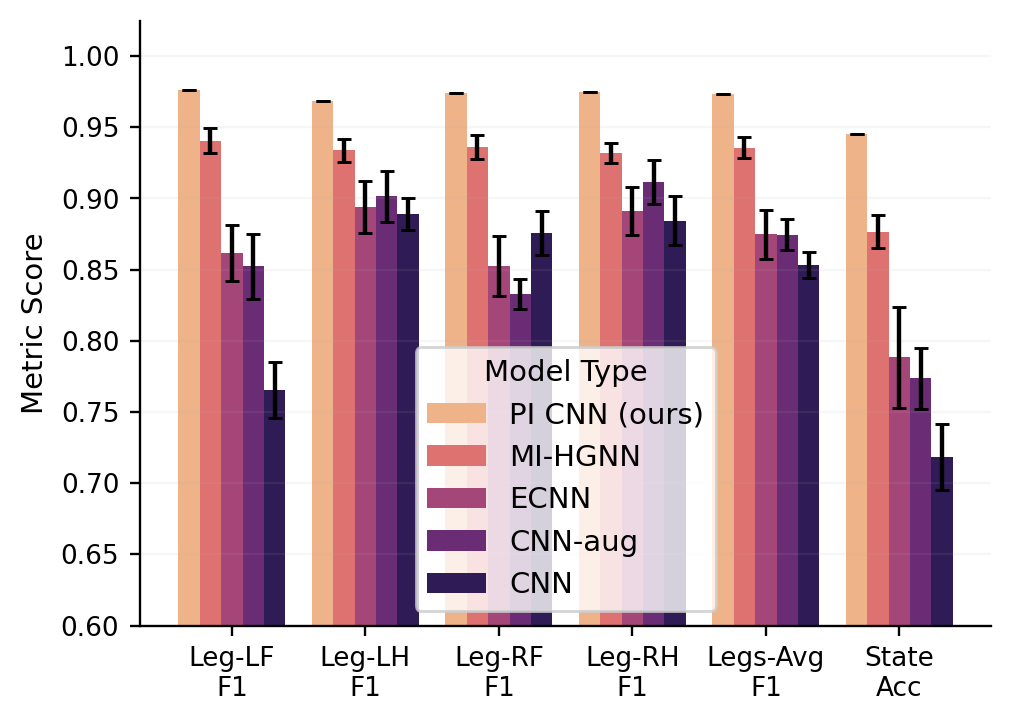} 
  \caption{Contact detection results on the real-world Mini-
Cheetah contact dataset \cite{Lin2021LeggedRS}: classification performance
of five models on the unseen test set, trained with the entire
training set. The mean and standard deviation across 8 runs
are reported and compared with MI-HGNN \cite{Butterfield2024MIHGNNMH}, ECNN \cite{Apraez2023OnDS}, CNN-aug \cite{Apraez2023OnDS} and CNN \cite{Lin2021LeggedRS}.}
  \label{fig:results_all}
\end{figure}

We also evaluated our approach using the same metrics as \cite{Apraez2023OnDS} \cite{Butterfield2024MIHGNNMH}, namely the foot-wise binary F1-score, the averaged F1-score (mean of the four leg-specific F1-scores), and the contact state accuracy over 16 possible states. The latter is a stricter metric than averaged accuracy, as a prediction is counted as correct only when the contact state of all four feet is simultaneously correctly predicted. To compute this metric for our model, the foot-wise predictions were converted into a 16-state representation. In addition, the real-world dataset provided by \cite{Lin2021LeggedRS} was transformed into \textsc{PIs} using a sliding window of size $w=10$, consistent with the results reported in Tab.~\ref{tab:multi_dataset_results}. Unlike the original setup, however, our model was trained on the merged training and validation splits (treated as a single training set) and evaluated on the merged test splits. The experimental results, presented in Fig.~\ref{fig:results_all}, demonstrate that our method outperforms all state-of-the-art data-driven methods (MI-HGNN \cite{Butterfield2024MIHGNNMH}, ECNN \cite{Apraez2023OnDS}, CNN-aug \cite{Apraez2023OnDS}, and CNN \cite{Lin2021LeggedRS}) in the Legs-Avg F1 (from 0.9356 of MI-HGNN to 0.9732) and overall model accuracy (from 87.7\% of MI-HGNN to 94.5\%) with smaller standard deviation. These results show the potential of this representation to achieve higher accuracy with a less complex NN architecture compared to other methods~\cite{Apraez2023OnDS},~\cite{Butterfield2024MIHGNNMH}.

\subsection{Simulation Results}
\begin{figure}[h!]

  \subfigure[Uneven Terrain]{%
    \includegraphics[width=.31\linewidth]{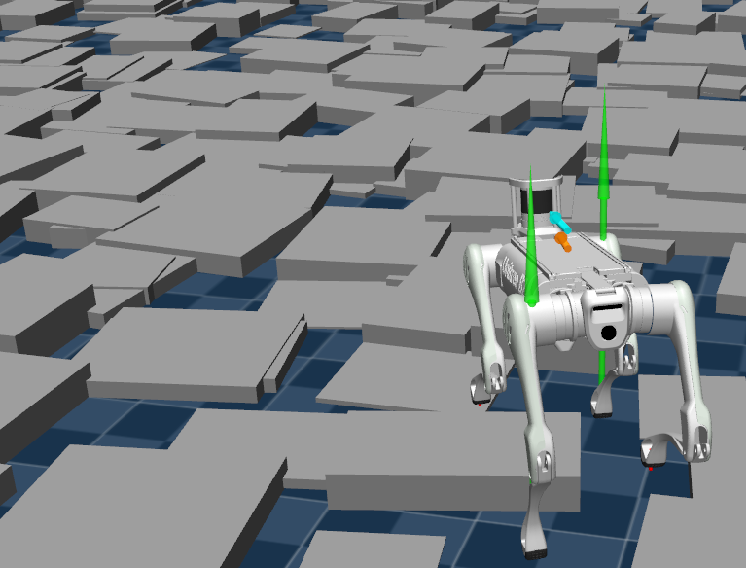}
    \label{fig:ds1}
  } 
  \subfigure[Flat Terrain]{%
    \includegraphics[width=.31\linewidth]{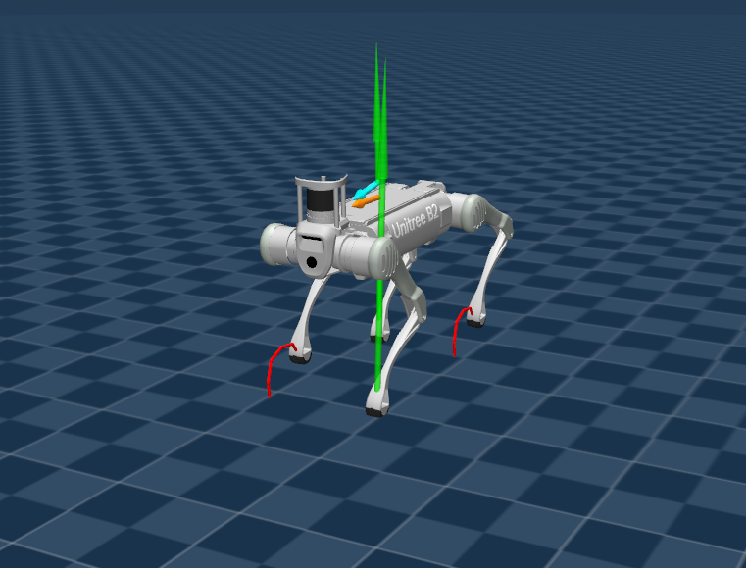}
    \label{fig:ds2} 
  }
  \subfigure[Stairs Terrain]{%
    \includegraphics[width=.31\linewidth]{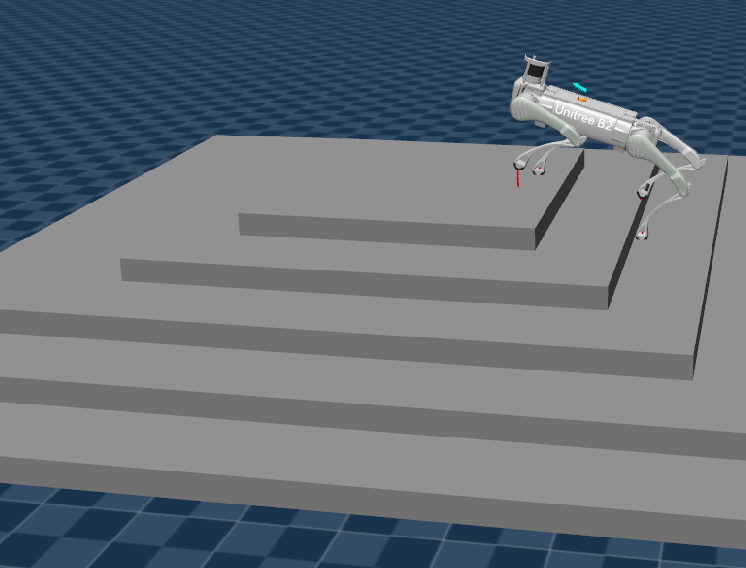}
    \label{fig:ds3}
  } 
  \caption{Mujoco Simulated Environment.} 
  \label{fig:Mujoco}
\end{figure}

To further evaluate our \textsc{PI} formulation, we generated a dataset in the MuJoCo simulator \cite{Todorov2012MuJoCoAP} using a 60kg Unitree B2 quadruped robot. The dataset includes three environments (flat, uneven, and pyramid), two gaits (trot and crawl), and both stable ($\mu$=1.0) and slippery ($\mu$=0.2) terrain conditions, where $\mu$ is the friction coefficient. The uneven and pyramid environments were randomized in relation to the boxes positions and the height of the pyramid. The robot traversed these environments under a gradient-based MPC controller developed in \cite{turrisi2024sampling}, with trunk linear velocities ranging from –1 to 1 m/s and angular velocities from –0.3 to 0.3 rad/s. Proprioceptive signals were recorded, including joint positions, joint velocities, foot positions, foot velocities, IMU signals and GRFs with Gaussian noise added to better approximate real-world sensing conditions. Contact estimation labels were directly extracted from the simulator and used as ground truth for training and evaluation. In total, the dataset comprises more than 1.5M synced data samples, organized by robot gait and environment stability. Each subset was further divided into training, validation, and test splits to enable fair comparison between our \textsc{PI} formulation, the approach of \cite{Lin2021LeggedRS}, and a baseline method based on GRF threshold \cite{Nistic2025MUSEAR,Fink2020ProprioceptiveSF}.  Figure \ref{fig:Mujoco} illustrates examples of scenes from the simulated dataset across the different environments.

\begin{table}[h!]
\centering
\begin{adjustbox}{width=0.48\textwidth}
\begin{tabular}{l c c c c c c c c c}
\toprule
\textbf{\Large Test Dataset} & \textbf{\Large Method} & \textbf{\Large Trained} &\textbf{\Large Acc(\%)} & \textbf{\Large F1} & \textbf{\Large LF(\%)} & \textbf{\Large RH (\%)} & \textbf{\Large RF (\%)} & \textbf{\Large LH (\%)} & \textbf{\Large Leg Avg (\%)} \\
\midrule
\multirow{4}{*}{\textbf{\LARGE Trot - Stable}} 
& \LARGE \textsc{PI} CNN  & \LARGE Stable &  \textbf{\LARGE 99.46} &  \textbf{\LARGE 0.9600} &  \LARGE 99.86 &  \LARGE 99.86 & \LARGE 99.90 &  \LARGE 99.80 &  \LARGE 98.85 \\
\cmidrule(lr){2-10}
& \LARGE \textsc{PI} CNN  & \LARGE Fused & \textbf{\LARGE 99.39} & \textbf{\LARGE 0.9565} & \LARGE 99.80 & \LARGE 99.83 & \LARGE 99.88 & \LARGE 99.78 & \LARGE 99.82 \\
\cmidrule(lr){2-10}
& \LARGE Lin \textit{et al.} \cite{Lin2021LeggedRS} & \LARGE Stable & \LARGE 96.69 & \LARGE 0.5840 & \LARGE 99.00 & \LARGE 98.96 & \LARGE 99.22 & \LARGE 99.14 & \LARGE 99.08 \\
\cmidrule(lr){2-10}
& \LARGE Lin \textit{et al.} \cite{Lin2021LeggedRS} & \LARGE Fused & \LARGE 96.79 & \LARGE 0.7291 & \LARGE 98.99 & \LARGE 98.91 & \LARGE 98.88 & \LARGE 99.05 & \LARGE 98.96 \\
\cmidrule(lr){2-10}
& \LARGE GRF TH & \LARGE - & \LARGE - & \LARGE - & \LARGE 69.741 & \LARGE 69.58 & \LARGE 69.41 & \LARGE 69.62 & \LARGE 69.95 \\
\midrule
\multirow{4}{*}{\textbf{\LARGE Trot - Slippery}} 
& \LARGE \textsc{PI} CNN  & \LARGE Stable & \textbf{\LARGE 88.41} & \textbf{\LARGE 0.7165} & \LARGE 96.36 & \LARGE 96.09 & \LARGE 95.99 & \LARGE 95.50 & \LARGE 95.98 \\
\cmidrule(lr){2-10}
& \LARGE \textsc{PI} CNN  & \LARGE Fused & \textbf{\LARGE 96.16} & \textbf{\LARGE 0.8925} & \LARGE 98.96 & \LARGE 99.13 & \LARGE 98.85 & \LARGE 98.79 & \LARGE 98.93 \\
\cmidrule(lr){2-10}
& \LARGE Lin \textit{et al.} \cite{Lin2021LeggedRS} & \LARGE Stable & \LARGE 86.69 & \LARGE 0.3510 & \LARGE 94.31 & \LARGE 94.62 & \LARGE 95.67 & \LARGE 95.64 & \LARGE 95.06 \\
\cmidrule(lr){2-10}
& \LARGE Lin \textit{et al.} \cite{Lin2021LeggedRS} & \LARGE Fused & \LARGE 91.24 & \LARGE 0.5708 & \LARGE 96.85 & \LARGE 96.67 & \LARGE 97.71 & \LARGE 97.26 & \LARGE 97.12 \\
\cmidrule(lr){2-10}
& \LARGE GRF TH & \LARGE - & \LARGE - & \LARGE - & \LARGE 66.14 & \LARGE 65.75 & \LARGE 64.81 & \LARGE 65.22 & \LARGE 65.485 \\
\midrule
\multirow{4}{*}{\textbf{\LARGE Trot - Fused}} 
& \LARGE \textsc{PI} CNN  & \LARGE Stable & \textbf{\LARGE 93.77} & \textbf{\LARGE 0.7899} & \LARGE 98.05 & \LARGE 97.93 & \LARGE 97.89 & \LARGE 97.58 & \LARGE 97.86 \\
\cmidrule(lr){2-10}
& \LARGE \textsc{PI} CNN  & \Large Fused & \textbf{\LARGE 97.73} & \textbf{\LARGE 0.9103} & \LARGE 99.36 & \LARGE 99.47 & \LARGE 99.35 & \LARGE 99.27 & \LARGE 99.36 \\
\cmidrule(lr){2-10}
& \LARGE Lin \textit{et al.} \cite{Lin2021LeggedRS} & \LARGE Stable & \LARGE 91.45 & \LARGE 0.3893 & \LARGE 96.55 & \LARGE 96.74 & \LARGE 97.32 & \LARGE 97.16 & \LARGE 96.94 \\
\cmidrule(lr){2-10}
& \LARGE Lin \textit{et al.} \cite{Lin2021LeggedRS} & \LARGE Fused & \LARGE 93.61 & \LARGE 0.5923 & \LARGE 97.84 & \LARGE 97.73 & \LARGE 98.05 & \LARGE 98.12 & \LARGE 97.93 \\
\cmidrule(lr){2-10}
& \LARGE GRF TH & \LARGE - & \LARGE - & \LARGE - & \LARGE 67.93 & \LARGE 67.66 & \LARGE 67.10 & \LARGE 67.105 & \LARGE 67.53 \\
\midrule
\multirow{4}{*}{\textbf{\LARGE Crawl - Stable}} 
& \LARGE \textsc{PI} CNN  & \LARGE Stable & \textbf{\LARGE 98.85} & \textbf{\LARGE  0.9982} & \LARGE 99.76 &  \LARGE 99.71 & \LARGE 99.67 &  \LARGE 99.70 & \LARGE 99.71 \\
\cmidrule(lr){2-10}
& \LARGE \textsc{PI} CNN  & \LARGE Fused & \textbf{\LARGE 98.87} & \textbf{\LARGE 0.9462} & \LARGE 99.73 & \LARGE 99.73 & \LARGE 99.68 & \LARGE 99.71 & \LARGE 99.71 \\
\cmidrule(lr){2-10}
& \LARGE Lin \textit{et al.} \cite{Lin2021LeggedRS} & \LARGE Stable & \LARGE 96.73 & \LARGE 0.7347 & \LARGE 99.03 & \LARGE 99.03 & \LARGE 99.28 & \LARGE 99.19 & \LARGE 99.13 \\
\cmidrule(lr){2-10}
& \LARGE Lin \textit{et al.} \cite{Lin2021LeggedRS} & \LARGE Fused & \LARGE 96.17 & \LARGE 0.7595 & \LARGE 99.01 & \LARGE 98.66 & \LARGE 99.20 & \LARGE 99.11 & \LARGE 99.00 \\
\cmidrule(lr){2-10}
& \LARGE GRF TH & \LARGE - & \LARGE - & \LARGE - & \LARGE 82.45 & \LARGE 82.81 & \LARGE 81.62 & \LARGE 81.49 & \LARGE 82.09 \\
\midrule
\multirow{4}{*}{\textbf{\LARGE Crawl - Slippery}} 
& \LARGE \textsc{PI} CNN  & \LARGE Stable & \textbf{\LARGE 83.48} & \textbf{\LARGE 0.6308} & \LARGE 95.64 & \LARGE 96.15 & \LARGE 94.39 & \LARGE 94.07 & \LARGE 95.06 \\
\cmidrule(lr){2-10}
& \LARGE \textsc{PI} CNN  & \LARGE Fused & \textbf{\LARGE 93.18} & \textbf{\LARGE 0.8575} & \LARGE 98.62 & \LARGE 98.36 & \LARGE 98.11 & \LARGE 97.57 & \LARGE 98.16 \\
\cmidrule(lr){2-10}
& \LARGE Lin \textit{et al.} \cite{Lin2021LeggedRS} & \LARGE Stable & \LARGE 78.48 & \LARGE 0.4098 & \LARGE 92.89 & \LARGE 92.14 & \LARGE 94.94 & \LARGE 93.78 & \LARGE 93.44 \\
\cmidrule(lr){2-10}
& \LARGE Lin \textit{et al.} \cite{Lin2021LeggedRS} & \LARGE Fused & \LARGE 84.34 & \LARGE 0.6021 & \LARGE 94.75 & \LARGE 95.10 & \LARGE 96.40 & \LARGE 96.21 & \LARGE 95.61 \\
\cmidrule(lr){2-10}
& \LARGE GRF TH & \LARGE - & \LARGE - & \LARGE - & \LARGE 77.54 & \LARGE 78.82 & \LARGE 74.28 & \LARGE 74.09 & \LARGE 76.18 \\
\midrule 
\multirow{4}{*}{\textbf{\LARGE Crawl - Fused}} 
& \LARGE \textsc{PI} CNN  & \LARGE Stable & \textbf{\LARGE 94.26} & \textbf{\LARGE 0.7685} & \LARGE 98.52 & \LARGE 98.64 & \LARGE 98.10 & \LARGE 98.03 & \LARGE 98.32 \\
\cmidrule(lr){2-10}
& \LARGE \textsc{PI} CNN  & \LARGE Fused & \textbf{\LARGE 97.17} & \textbf{\LARGE 0.8934} & \LARGE 99.40 & \LARGE 99.32 & \LARGE 99.21 & \LARGE 99.08 & \LARGE 99.25 \\
\cmidrule(lr){2-10}
& \LARGE Lin \textit{et al.} \cite{Lin2021LeggedRS} & \LARGE Stable & \LARGE 90.24 & \LARGE 0.5374 & \LARGE 96.71 & \LARGE 96.54 & \LARGE 97.75 & \LARGE 97.49 & \LARGE 97.12 \\
\cmidrule(lr){2-10}
& \LARGE Lin \textit{et al.} \cite{Lin2021LeggedRS} & \LARGE Fused & \LARGE 92.79 & \LARGE 0.6840 & \LARGE 97.55 & \LARGE 97.71 & \LARGE 98.38 & \LARGE 98.34 & \LARGE 97.99 \\
\cmidrule(lr){2-10}
& \LARGE GRF TH & \LARGE - & \LARGE - & \LARGE - & \LARGE 79.96 & \LARGE 80.87 & \LARGE 78.18 & \LARGE 77.71 & \LARGE 79.18 \\

\bottomrule
\end{tabular}
\end{adjustbox}
\caption{Contact Estimation comparison across datasets from Mujoco. Stable Data is the environment without slippery terrain, and Fused Data is the combined datasets of slippery and stable terrains.}
\label{tab:mujoco_dataset_results}
\end{table}
Table~\ref{tab:mujoco_dataset_results} presents a comparison among three approaches: our \textsc{PI}-based CNN, the method of Lin et al. \cite{Lin2021LeggedRS}, and a threshold-based GRF baseline. The MuJoCo dataset is used with the same window sizes as described in Section~\ref{sec:real_world}. Both models were trained on the stable environment and evaluated on three test sets: pure stable, slippery, and a fused (stable + slippery) dataset. The contact threshold was obtained by sweeping candidate values over the dataset’s GRF in the Z axis and selecting the value that maximized agreement with the ground-truth contact labels. The \textsc{PI}-CNN consistently outperformed the baseline across all test conditions, including the slippery environment, despite never being exposed to slippery data during training. These results highlight that \textsc{PI} provides a richer and more discriminative feature representation, enabling superior learning compared to directly using raw proprioceptive signals.

\subsection{Implementation Details}
All tests were performed on a computer with Intel i9 CPU with 64 GB of RAM and NVIDIA RTX3090 GPU running Ubuntu 20.04 LTS Linux. A dropout rate of 0.3 was applied in the fully connected layers as an additional form of regularization. Training was carried out using Cross-Entropy loss for multi-class classification, optimized with adamW \cite{Loshchilov2017DecoupledWD}, initial learning rate of $1 \cdot 10^{-4}$, trained through 40 epochs with batch size of 64. To improve convergence, a ReduceLROnPlateau scheduler was employed, which decreased the learning rate by a factor of 0.2 when the validation loss failed to improve for 3 consecutive epochs.

\section{Discussion}

Although \textsc{PI} encodings provide rich, learnable patterns, their generation is relatively slow (1.5 ms per image for $w=10$) due to the current CPU-based Python implementation. Increasing image resolution can provide richer feature content and improve accuracy, but it also increases encoding cost. In our experiments, the model achieved a mean GPU inference time of 8.5 ms ($\approx 120 Hz$). We therefore expect that a multi-threaded C++ implementation will substantially reduce generation time, making the formulation more suitable for real-time robotic systems. Beyond runtime, \textsc{PI}s derived from IMU signals exhibit Brownian-walk-like drift, which over long experiments leads to saturated, repetitive patterns. Finally, the kernel functions used to encode each PI component were heuristically tuned for contact estimation, suggesting that a more systematic study is needed to generalize the encoding to other tasks and robotic platforms.

\section{CONCLUSIONS}
This paper presented a novel formulation of proprioceptive signals into an image representation. Each image representation encodes four important types of temporal information related to dynamic measurement behavior and is concatenated based on the morphology of the quadruped robot. The \textsc{PI} was evaluated on the contact estimation problem, using a simple CNN architecture trained on both simulated and real-world data, achieving the best results in comparison with state-of-the-art methods. The \textsc{PI} is the first step toward deeper representations of robot sensing. For future work, we expect that the \textsc{PI} can be integrated into other learning structures such as GNN similar to \cite{Butterfield2024MIHGNNMH} and also applied as input for reinforcement learning techniques. Moreover, combining a \textsc{PI} with exteroceptive sensors, such as cameras, could yield a unified representation of both the robot’s internal state and its external environment, further enhancing perception and state learning. Finally, we aim to explore the applicability of \textsc{PI} in other robotic platforms, including humanoids and manipulators.







\bibliographystyle{IEEEtran}
\bibliography{main}


\end{document}